\pdfoutput=1

\documentclass[11pt]{article}

\usepackage{acl}

\usepackage{times}
\usepackage{latexsym}

\usepackage[T1]{fontenc}

\usepackage[utf8]{inputenc}

\usepackage{microtype}

\usepackage{inconsolata}
\usepackage{graphicx}
\usepackage{subcaption}
\usepackage{booktabs}
\usepackage{multirow}
\usepackage{enumitem}
\usepackage{amsmath,amssymb}
\usepackage{color}
\usepackage{soul}
\usepackage{natbib}
\usepackage{mathtools}

\usepackage{array}
\usepackage{amsthm}
\usepackage{bbold}
\usepackage{cleveref}

\title{N-Gram Nearest Neighbor Machine Translation}

\author{
  Rui Lv$^{1}$\thanks{~ Equal contribution. Work done when Rui Lv is interning at Microsoft Research Asia.}\ \ , Junliang Guo$^2$\footnotemark[1]\ \ , Rui Wang$^2$, Xu Tan$^2$, Qi Liu$^{1}$, Tao Qin$^{2}$  \\
  $^1$University of Science and Technology of China  
  $^2$Microsoft Research Asia \\
  $^1$\texttt{lvrui2018@mail.ustc.edu.cn},\; \texttt{qiliuql@ustc.edu.cn}\\ 
  $^2$\texttt{\{junliangguo, ruiwa, xuta, taoqin\}@microsoft.com}\\
}

\begin{document}
\maketitle
\begin{abstract}
Nearest neighbor machine translation 
augments
the Autoregressive Translation~(AT) with $k$-nearest-neighbor retrieval,
by comparing the similarity between the token-level context representations of the target tokens in the query and the datastore.
However, the token-level representation may introduce noise when translating ambiguous words, or fail to provide accurate retrieval results when the representation generated by the model contains indistinguishable context information, e.g., Non-Autoregressive Translation~(NAT) models.
In this paper, we propose a novel $n$-gram nearest neighbor retrieval method that is model agnostic and applicable to both AT and NAT models. 
Specifically, 
we concatenate the adjacent $n$-gram hidden representations 
as the key, while the tuple of corresponding target tokens is the value. 
In inference, we propose tailored decoding algorithms for AT and NAT models respectively.
We demonstrate that the proposed method consistently outperforms the token-level method on both AT and NAT models as well on general as on domain adaptation translation tasks. On domain adaptation, the proposed method brings $1.03$ and $2.76$ improvements regarding the average BLEU score on AT and NAT models respectively.
\end{abstract}

\section{Introduction}
\label{sec:intro}
Retrieval-based methods augment neural machine translation models~\citep{bahdanau2014neural,vaswani2017attention} with external memory to provide more scalable and adaptable translation ability, such as dealing with out-of-domain or rare word translations~\citep{gu2018search,zhang2018guiding,cao2018encoding,bapna2019non,khandelwal2020nearest}. Among them, the recently proposed $k$-Nearest-Neighbor Machine Translation~($k$NN-MT) model~\citep{khandelwal2020nearest} achieves promising results on domain adaptation without tuning model parameters.
Specifically, $k$NN-MT utilizes a pre-trained NMT model to construct a datastore that contains context representations as well as corresponding target tokens in the training set, and in inference, relevant tokens that may help the translation of the current token are retrieved by taking the current context representation as the query.
This non-parametric way equips the model with fast adaptation and generalization abilities and has been widely adopted
in text, speech and image generations~\citep{khandelwal2019generalization,du2022non,ashual2022knn}.

However, as the token-level representation sometimes fails to deal with ambiguous context information, the retrieved neighbors may contain noise that affects the accuracy and robustness of translation results~\citep{zheng2021adaptive}. In addition, as an important 
alternative to traditional Autoregressive Translation~(AT) models, Non-Autoregressive Translation~(NAT) models~\citep{gu2018non} 
achieve significant acceleration regarding
the decoding speed by removing the conditional dependency and generating all target tokens in parallel, 
which also leads to inferior translation accuracy due to the lack of context information. 
Previous works have shown that the hidden representations encoded by NAT models are of low quality, for example, sometimes the representations of adjacent positions are indistinguishable~\citep{wang2019non,guo2019non}. 
In this scenario, 
directly utilizing token-level retrieval can only achieve sub-optimal results
as low quality context representations cannot retrieve accurate values.

To deal with the above problems, in this paper, we propose a novel $n$-gram~\citep{joachims1998text} based nearest neighbor~($n$-$k$NN) retrieval method. Specifically, instead of taking a single hidden representation and its target token as the key and value, we incorporate the hidden representations of adjacent $n$ tokens as the key and the tuple of their corresponding target tokens as the value. By taking consecutive tokens into consideration, the datastore provides additional phrase-level information which is more accurate and informative.

During inference, through nearest neighbor retrieval, each position in the $n$-gram group will obtain a $k$NN prediction distribution, which is then augmented with the model prediction to determine the final result.
However, it is non-trivial to make the retrieved $n$-gram information fully utilized. For AT models, due to the token by token prediction manner, $n$-gram will degenerate to unigram as only the prediction of the current step can be directly modified. For NAT models, we empirically find that the $n$-gram retrieval brings marginal improvements with the default decoding algorithm.
To deal with the challenges, 
we design different decoding algorithms for AT and NAT models. 
We integrate the retrieved $n$-gram values into the beam search decoding of the AT model to provide it an opportunity to re-weighting previous translations.
For NAT models, we propose a two-pass decoding algorithm where the model is forwarded twice to generate context representations with better quality.
We verify that the proposed $n$-gram retrieval and 
corresponding decoding algorithms
are both essential for the success of the whole process.

We evaluate the proposed method on AT and NAT models on both general and multi-domain machine translation tasks. The proposed $n$-$k$NN method consistently outperforms the vanilla $k$NN method, specifically, on domain adaptation tasks, $n$-$k$NN achieves $1.03$ and $2.76$ BLEU scores improvements on average with AT and NAT models respectively. On general translation tasks, $n$-$k$NN with the NAT model also brings $0.48$ and $1.10$ BLEU scores improvements on WMT14 German-English and WMT16 English-Romanian translation tasks.

\section{Related Work}

\subsection{Nearest Neighbor Machine Translation}
The $k$-Nearest-Neighbor Machine Translation~($k$NN-MT)~\citep{khandelwal2020nearest} generally contains two steps: datastore creation and $k$NN prediction. Specifically, given a pre-trained autoregressive machine translation model $f(\cdot)$, the training set that contains bilingual pairs $(x,y) \in (\mathcal{X}, \mathcal{Y})$, we denote the hidden representation when predicting the $t$-th target token $y_t$ as $h_t=f(x, y_{<t})$\footnote{We use the hidden representation before the final softmax layer as $h$, mainly follow the setting in previous works~\citep{khandelwal2020nearest}}. Then the datastore is constructed by a single forward pass over each target token in the training set:
\begin{equation*}
(\mathcal{K}, \mathcal{V}) = \bigcup_{(x, y) \in (\mathcal{X}, \mathcal{Y})} \{(h_t, y_t), \forall t \in [1, T_y] \},
\label{equ:datastore}
\end{equation*}
where $T_y$ indicates the length of target sentence $y$, the context representation $h_t$ and the target token $y_t$ are taken as the key and value respectively.

During inference, at each time-step $t$, the AT model generates the current context representation $\hat{h}_t=f(x, \hat{y}_{<t})$ conditioned on the source sentence $x$ and target tokens generated in previous steps $\hat{y}_{<t}$, which is utilized to query the datastore and retrieve $k$ nearest neighbors $N = \{(h_i, v_i), \forall i \in [1,k] \}$. The $k$NN prediction probability is then calculated by the distance between the query and retrieved keys,
\begin{equation}
p^{\textrm{kNN}}_t(y_t |x, \hat{y}_{<t}) \propto 
\sum_{(h_i, v_i) \in N} \mathbb{1}_{y_t = v_i} \exp (\frac{-d(h_i, \hat{h}_t)}{\tau}),
\label{equ:knn-prob}
\end{equation}
where $d(\cdot)$ indicates the $l_2$ distance and $\tau$ the temperature. Finally, the prediction probability of $y_t$ is calculated by interpolating the $k$NN prediction and AT model prediction with a hyper-parameter $\lambda$:
\begin{align}
\label{equ:prob_ip}
p(y_t|x, \hat{y}_{<t}) &= \lambda \ p^{\textrm{kNN}}_t(y_t|x, \hat{y}_{<t}) \\
& + (1-\lambda) \ p^{\textrm{AT}}_{t} (y_t|x, \hat{y}_{<t}). \nonumber
\end{align}
Recently, some followup works that improve the accuracy~\citep{jiang2021learning}, the generality~\citep{zheng2021adaptive}, the robustness ~\citep{jiang2022towards},  and the efficiency~\citep{wang2022efficient,alon2022neuro,meng2021fast,martins2022chunk,yang-etal-2022-nearest} of vanilla $k$NN-MT are proposed.

\subsection{Autoregressive and Non-Autoregressive Machine Translation}
\label{sec:rw_nat}

Traditional Autoregressive Translation~(AT) models generate the target translation word-by-word, i.e., given the source sentence $x=(x_1, x_2, ..., x_{T_{x}})$, the target $y=(y_1, y_2, ..., y_{T_{y}})$ is modeled as
$p(y|x)=\prod_{t=1}^{T_{y}}p(y_{t}|y_{<t},x)$, where $y_{<t}$ indicates the target tokens generated in previous timesteps, $T_{x}$ and $T_{y}$ indicate the length of source and target respectively. Non-Autoregressive Translation~(NAT) models break the conditional dependency and generate all tokens in parallel, $p(y|x)=p(T_{y}|x) \cdot \prod_{t=1}^{T_{y}}p(y_{t}|x)$, where $p(T_{y}|x)$ models the target length prediction. To achieve that, the decoder input of NAT models is usually a copy of source representations or an empty sequence constructed with \texttt{[UNK]} or \texttt{[MASK]} tokens~\citep{qian2020glancing,ghazvininejad2019mask}, which leads to the low quality of hidden representations and performance degradation.
The vanilla $k$NN method only considers the autoregressive model, and does not work for the non-autoregressive model according to our experiments. We propose the $n$-$k$NN method which performs well on both AT and NAT models.

\subsection{N-Gram Methods in Machine Translation}
$N$-gram based methods have been widely utilized in natural language processing tasks, including language modeling~\citep{brown1992class}, text classification~\citep{cavnar1994n} and scoring~\citep{kondrak2005n}, statistical machine translation~\citep{marino2006n} and neural machine translation~\citep{shao2018greedy}, etc.
In NAT, some works~\citep{shao2020minimizing,guo2020jointly} have utilized $n$-gram based loss functions to achieve better results. Different from them, our method is non-parametric and
does not change the model parameters.

It is worth noting that~\citep{bapna2019non} also introduces $n$-gram based retrieval to help the adaptation of NMT. Our work differs in
two points: 1) their adaption is achieved by retrieving and attending similar source sentences in the sentence level, which requires additional training and may add noise;
2) we concentrate on the non-parametric adaptation of both AT and NAT models through $n$-gram level retrieval, which has not been studied in the literature to the best of our knowledge.

\begin{figure*}[tb]
\centering
\includegraphics[width=0.8\linewidth]{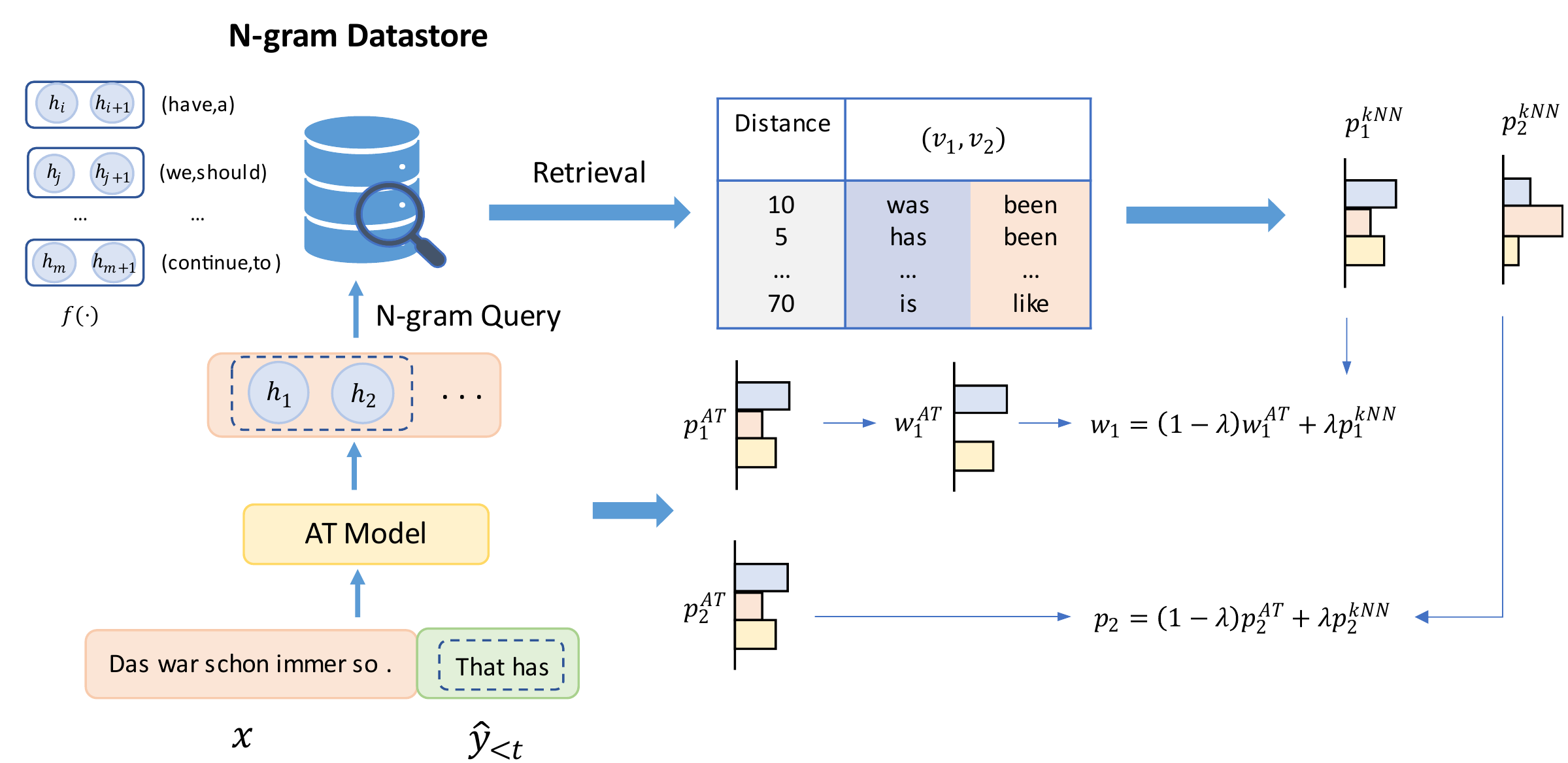}
\vspace{-5pt}
\caption{An illustration of $n$-$k$NN on AT with $n=2$ as an example. 
$h_1$ and $h_2$ indicate the hidden representation of ``That'' and ``has'' respectively and we incorporate them with mean pooling $f(\cdot)$, while $p^{AT}_1$ and $p^{AT}_2$ indicate the 
prediction distributions of the two steps, $w_1$ is the beam weight of the first token. The $n$-gram retrieved values are utilized to update the beam weight as well as help the current prediction. }
\label{fig:nknn-at}
\vspace{-10pt}
\end{figure*}

\section{Methodology}

We will introduce the proposed $n$-gram $k$ Nearest Neighbor~($n$-$k$NN) method in this section, including the $n$-gram datastore creation and retrieval augmented decoding for AT and NAT models.

\subsection{N-Gram Datastore Creation}
\label{sec:datastore-creation}

Given a translation pair $(x,y)\in (\mathcal{X}, \mathcal{Y})$ in the training set, for target tokens $y=(y_1,y_2, ..., y_{T_y})$, the corresponding context representations are denoted as $h=(h_1,h_2, ..., h_{T_y})$. 
We denote the $n$-gram representations as $g(h_t)=(h_{t-n+1}, ..., h_{t})$, then
we can collect all possible $n$-gram representations in $y$ as:
\begin{equation}
\mathcal{K} = \{ f(g(h_{t})), n \leq t \leq T_y \}, \nonumber
\end{equation}
where $f(\cdot)$ indicates the integration function of $n$-gram representations, with several common choices including concatenation, mean pooling, etc. We utilize mean pooling in our paper, i.e.,
\begin{equation}
f(g(h_{t})) = \frac{1}{n} \sum_{i=1}^{n} h_{t-i+1}, \nonumber
\label{interpolate-f}
\end{equation}
The corresponding $n$-gram target tokens can be denoted as:
\begin{equation}
\mathcal{V} = \{ g(y_t), n \leq t \leq T_y \} \}, \nonumber
\end{equation}
where $g(y_t) = (y_{t-n+1}, ..., y_{t})$, i.e., each key is a tuple with $n$ tokens.
Then $(\mathcal{K}, \mathcal{V})$ is utilized to create the $n$-gram datastore.
It is worth noting that the scale of $n$-gram datastore is similar to the original unigram counterpart, which contains $(T_y-n+1)$ and $T_y$ key-value pairs for each sentence respectively, and the dimension of key vectors is also the same with mean pooling integration.
The proposed $n$-gram level datastore contains phrase-level information, which helps the model better distinguish ambiguous token representations.

After the datastore is created, another important procedure is retrieving it while inference. 
Given an $n$-gram query $q_t=f(g(\hat{h}_{t}))$ generated by the model, we retrieve its $k$ nearest neighbors from the datastore, denoted as
\begin{equation}
    N_t=\{(f(g(h^i_t)), g(v^i_t)), \forall i\in [1,k] \},
\label{equ:neighbor-n-gram}
\end{equation}
which contains $k$ key representations and value tokens\footnote{Target tokens in the datastore are represented as $v$, while $y$ in other cases.}. As each position will be covered by multiple $n$-gram objects, e.g., $h_2$ exists in consecutive $2$-grams
$(h_1,h_2)$ and $(h_2,h_3)$,
how to fully utilize the information contained in retrieved $n$-grams is a non-trivial problem.
We propose tailored retrieval algorithms for AT and NAT models respectively considering their different nature in inference, and describe them in Section~\ref{sec:n-gram-at} and Section~\ref{sec:n-gram-nat}.

\subsection{N-Gram Augmented Beam Search for AT}
\label{sec:n-gram-at}

Beam search~\cite{graves2012sequence} has become the de facto approach for searching optimal decoding results in autoregressive models. In each translation step $t$, it maintains $b$ topmost translations $\hat{y}_t$\footnote{We omit the beam index in the following as the computation for each beam is the same.} with corresponding hidden representations $\hat{h}_{t}$.

For the current step $t$ where the prediction has not been made yet, we use the query $\hat{h}_t$, the retrieved key and value in the current step $h_t$ and $v_t$ to calculate the $k$NN prediction, and integrate it with the original model prediction following Eqn~(\ref{equ:prob_ip}). 
As for the previous steps in the $n$-gram,  
due to the left-to-right generation manner of AT models, 
their predictions 
have been determined and thus
the retrieved $n$-gram results for the previous steps cannot directly change their predictions. Instead, we use them to modify the beam weights 
of previous steps, which are calculated by accumulating the translation probability of each predicted token so far.
Specifically, 
we cache the prediction probability of the previous step $p^{\textrm{AT}}_{t-1}(y_{t-1}|x, \hat{y}_{<t-1})$ in each decoding step.
Given the retrieved $n$-gram neighbors $N_t$,
the $k$NN distribution for values retrieved at different steps can be denoted as:
\begin{equation} 
\label{equ:at-ngram-knn}
 p^{\textrm{n-kNN}}_{t-i}  \propto 
 \sum_{(f(g(h_t)), g(v_t))  \in N_t}  \mathbb{1}_{\hat{y}_{t-i} = v_{t-i}} \exp (\frac{-d_t}{\tau}), 
\end{equation}
where $i \in [0,n-1]$ denotes position offsets in the $n$-gram. 
$\hat{y}_{t-i}$ and $v_{t-i}$ indicate the current prediction and its value in the retrieved $n$-grams respectively, 
and we use the same weight for values in the same $n$-gram since $d_t=d(f(g(h_t)), q_t)$ is equivalent for all positions on $q_t$.
Therefore, Eqn~(\ref{equ:at-ngram-knn}) measures the $k$NN distribution of retrieved tokens that are at both the previous and current prediction steps.
For the $k$NN prediction of the previous step, we use it to update the beam score to re-weight the predictions, denoted as:
\begin{equation} 
\label{equ:update-beam-weight}
w(y_{t-i}|y_{<t-i}) = (1- \lambda) w(y_{t-i}|y_{<t-i}) + \lambda p^{\textrm{n-kNN}}_{t-i}, \nonumber
\end{equation}
where $w(y_{t-i}|y_{<t-i})$ denotes the beam weight of the $(t-i)$-th position.
And for the current step, 
the $k$NN prediction is augmented with the model prediction in a similar way as Eqn~(\ref{equ:prob_ip}) for the final probability:
\begin{equation}
p(y_t|x, y_{<t}) = (1- \lambda) p^{\textrm{AT}}_{t}(y_t|x, y_{<t}) + \lambda p^{\textrm{n-kNN}}_{t}. \nonumber
\label{equ:n-knn-at-final}
\end{equation}
As a result, on one hand, the proposed $n$-gram $k$NN method helps the model prediction of the current step considering phrase-level information through traditional prediction interpolating; on the other hand, it also enables the AT model to refine
previous decisions by changing their weights according to the retrieved results.
We provide an illustration of the proposed $n$-gram augmented beam search process in Figure~\ref{fig:nknn-at}.

\subsection{Two-Pass Decoding Algorithm for NAT}
\label{sec:n-gram-nat}

\begin{figure*}[tb]
\centering
\includegraphics[width=0.8\linewidth]{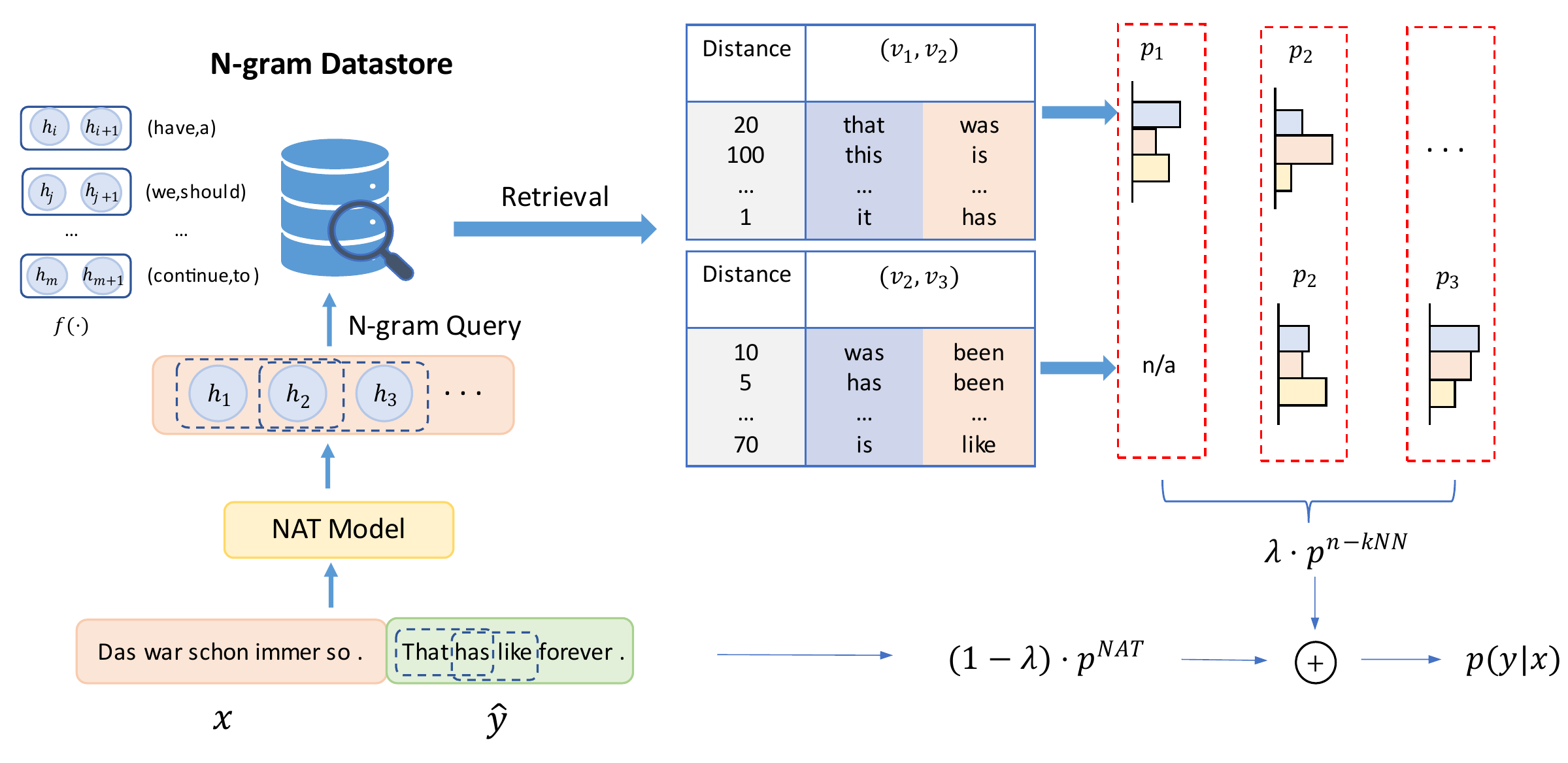}
\vspace{-5pt}
\caption{An illustration of $n$-$k$NN on NAT with $n=2$ as an example. $h_1, h_2, h_3$ represent the hidden representations for ``That'', ``has'' and ``like'' respectively.
And $f(g(h_1, h_2))$ and $f(g(h_2, h_3))$ are taken as two queries on step $1$ and $2$ respectively. The $k$NN distributions regrading the same position in different retrieved values~($p_2$ in this example) are augmented to perform joint prediction.}
\label{fig:nknn-nat}
\vspace{-10pt}
\end{figure*}

As introduced in Section~\ref{sec:rw_nat}, the decoder input of NAT model contains little target information, which makes the generated hidden representation of poor quality and indistinguishable. To alleviate the problem, we design a two-pass decoding algorithm equipped with $n$-gram retrieval for NAT,
which only happens in inference and does not change the training of the model. 
Specifically, given the source sentence $x$, the NAT model generates a candidate translation in the first pass:
\begin{equation}
p(\hat{y}|x)=p(T_{y}|x) \cdot \prod_{t=1}^{T_{y}}p(\hat{y}_{t}|x),
\end{equation}
where $\hat{y}=(\hat{y}_1,\hat{y}_2, ..., \hat{y}_{T_y})$ indicates the translation results in the first pass, which is fed into the model as the decoder input of the second pass:
\begin{equation}
p(y|x, \hat{y})= \prod_{t=1}^{T_{y}}p(y|\hat{y}_{t},x).
\end{equation}
The length prediction is omitted in the second pass as it has been determined in the first pass. 
We take the hidden representation in the second pass
as the key when creating the datastore, as well as the query utilized in retrieval.

Different from AT, the NAT model generates all tokens in parallel. Then given the query at the $t$-th position $q_t=f(g(\hat{h}_t))$, the retrieved neighbors $N_t$ defined in Eqn~(\ref{equ:neighbor-n-gram}),
the $k$NN prediction for each element is calculated as:
\begin{align}
&p_{\textrm{t}}^{\textrm{n-kNN}}(y_{t-i} |x,  \hat{y}, N_t) \propto \\ \nonumber
&\sum_{(f(g(h_t)), g(v_t)) \in N_t}  \mathbb{1}_{y_{t-i} = v_{t-i}} \exp (\frac{-d_t}{\tau}),
\label{equ:ngram-knn-prob}
\end{align}
where $i \in [0,n-1]$ denotes position offsets in the $n$-gram, $v_{t-i}$ denotes the corresponding retrieved value, and $d_t=d(f(h_t), q_t)$ denotes the same weight for all values in the retrieved $n$-gram.
The $k$NN predictions regarding the $t$-th step in different $n$-grams are augmented to construct the overall $k$NN prediction:
\begin{equation}
p^{\textrm{n-kNN}}(y_t |x, \hat{y}) = \sum_{j=0}^{n-1} p_{\textrm{t}}^{\textrm{n-kNN}}(y_t |x, \hat{y}, N_{t+j}).
\label{equ:ngram-knn-final}
\end{equation}
Then the final prediction probability of the NAT model is calculated as the interpolation of the $n$-gram $k$NN prediction as well as the original prediction:
\begin{equation}
\label{equ:prob_ip_ngram_knn}
p(y_t|x) =  \lambda  p^{\textrm{n-kNN}}(y_t|x, \hat{y}) 
 + (1-\lambda)  p^{\textrm{NAT}} (y_t|x, \hat{y}).   \nonumber
\end{equation}
The illustration of utilizing $n$-$k$NN in NAT is shown in Figure~\ref{fig:nknn-nat}.

\subsection{Discussion}
\label{sec:discuss}
The proposed method differs from the vanilla token-level $k$NN in the following aspects.
1) By storing and retrieving via $n$-gram representation, the model is able to utilize the phrase-level information in the dataset, which is richer than the token information, to alleviate the inaccurate translation of ambiguous tokens. 
2) The $n$-gram retrieval makes the AT model able to re-weight the previous translations in each step, e.g., emphasizing the token that appears many times in retrieved $n$-grams, or reducing the importance of tokens that are rarely retrieved and probably incorrect. 
And on the NAT model, the proposed $n$-gram $k$NN can reduce the error rate of repetitive translations, a well-known problem of NAT models, with the help of $n$-gram representation~\citep{shao2020minimizing,guo2020jointly}.
4) As we introduced in Section~\ref{sec:datastore-creation}, 
the datastore constructed by the $n$-$k$NN method contains similar number of key-value pairs as the vanilla unigram $k$NN. And the dimension of query/key is also the same. Therefore, $n$-$k$NN has similar time and space computational as vanilla $k$NN.
5) The proposed method is orthogonal to recent works that improve the performance or efficiency of the vanilla $k$NN~\citep{zheng2021adaptive,jiang2021learning,he2021efficient,wang2022efficient}.

\begin{table*}[tb]
\centering
\caption{The SacreBLEU scores of the proposed $n$-$k$NN method and the baseline methods with AT and NAT models on the De-En multi-domain tasks.
"$\Uparrow$" indicate the improvement of the $n$-$k$NN over $k$NN is significant at the level of $0.01$. The Base AT model is the WMT19 de-en winner, while the Base NAT model is trained on WMT14 De-En by ourselves.
The baseline scores are reported by our own implementation of $k$NN. 
}
\vspace{-10pt}
\begin{tabular}{l|cccc|c}
\toprule
 \textbf{Models}   & \multicolumn{1}{c}{\textbf{IT}}& \multicolumn{1}{l}{\textbf{Koran}} & \multicolumn{1}{c}{\textbf{Law}} & \multicolumn{1}{c}{\textbf{Medical}} &  \multicolumn{1}{|c}{\textbf{Aver}} \\
\midrule
Base AT & $37.97$ & $16.90$ & $45.85$ & $40.49$ &  $35.30$ \\
$+$ $k$NN & $44.54$ & $20.38$ & $60.98$ & $54.22$ & $45.03$ \\
$+$ \textbf{$n$-$k$NN} & $\textbf{46.42}^\Uparrow$ & $\textbf{21.39}^\Uparrow$ & $\textbf{61.98}$ & $\textbf{55.27}^\Uparrow$ & $\textbf{46.26}^\Uparrow$ \\
\midrule
Base NAT & $30.60$ & $10.42$ & $33.89$ & $30.15$ & 26.26 \\
$+$ $k$NN & $30.59$ & $10.46$ & $33.85$ & $30.06$  &  26.24 \\
$+$ \textbf{$n$-$k$NN} & $\textbf{31.43}^\Uparrow$ & $\textbf{10.93}^\Uparrow$ &  $\textbf{37.77}^\Uparrow$ & $\textbf{35.95}^\Uparrow$ & $\textbf{29.02}^\Uparrow$ \\
\bottomrule
\end{tabular}
\label{tab:domain_bleu_results}
\vspace{-10pt}
\end{table*}

\begin{table}[tb]
\small
\centering
\caption{The tokenized BLEU scores of NAT model with the proposed $n$-$k$NN method and the baseline methods on IWSLT14 De-En, WMT14 De-En and WMT16 En-Ro general machine translation datasets.
"$\Uparrow$" indicates the improvement of the $n$-$k$NN over the $k$NN is significant at the level of 0.01.
}
\vspace{-10pt}
\begin{tabular}{l|cccc}
\toprule
 \multicolumn{1}{c|}{}   & \multicolumn{1}{c}{\textbf{IWSLT14}}& \multicolumn{1}{c}{\textbf{WMT14}} & \multicolumn{1}{c}{\textbf{WMT16}}    \\
  \textbf{Models} &\multicolumn{1}{c}{\textbf{De-En}}& \multicolumn{1}{c}{\textbf{De-En}} & \multicolumn{1}{c}{\textbf{En-Ro}} \\
\midrule
Base NAT & $29.65$ & $30.09$ & $29.02$ \\
$+$ $k$NN & $29.03$ & $30.04$ & $28.92$    \\
$+$\textbf{$n$-$k$NN} & $\textbf{30.72}^\Uparrow$& $\textbf{30.57}^\Uparrow$  & $\textbf{30.12}^\Uparrow$  \\
\bottomrule
\end{tabular}
\label{tab:general_bleu_results}
\vspace{-10pt}
\end{table}

\section{Experiments}

We evaluate the proposed method on both domain adaptation and general machine translation tasks, as well as AT and NAT models respectively. We start with the experimental setup, and please refer to Appendix~\ref{appen:exp} for more details.

\subsection{Experimental Setup}

\paragraph{Datasets}
For domain adaptation, we follow the previous works~\citep{khandelwal2020nearest,zheng2021adaptive,jiang2021learning} and utilize the multi-domain dataset released by~\citet{aharoni2020unsupervised}. We consider four domains named Koran, IT, Medical and Law. 
And
we use the widely adopted benchmark datasets for the general translation~\citep{gu2018non,guo2019non,wang2019non}, including IWSLT14 German-English~(IWSLT De-En)\footnote{\url{https://wit3.fbk.eu/}}, WMT14 German-English~(WMT De-En)\footnote{\url{https://www.statmt.org/wmt14/translation-task}}, and WMT16 English-Romanian~(WMT En-Ro)\footnote{\url{https://www.statmt.org/wmt16/translation-task}}.
To the best of our knowledge, we are the first to use the $k$-Nearest-Neighbor method on NAT, so we adopt sequence-level knowledge distillation~\citep{kim2016sequence} on WMT14 De-EN dataset to train the NAT model following the common practice~\citep{gu2018non}, and raw datasets for other tasks.
We tokenize the sentences by Moses\footnote{\url{https://github.com/moses-smt/mosesdecoder}}~\citep{koehn2007moses} and segment each word into subwords using Byte-Pair Encoding~(BPE)~\citep{sennrich2015neural}. 
We use the bpecodes provided by~\citet{ng2019facebook} for domain adaptation datasets, and generate from scratch for general datasets.

\paragraph{Model Configurations}
We utilize the Transformer model~\citep{vaswani2017attention} as the autoregressive backbone model, while GLAT~\citep{qian2020glancing} for the non-autoregressive model. 
For domain adaptation tasks, we use the winner model~\citep{ng2019facebook} of the WMT19 German-English translation task as the pre-trained AT model, following previous works~\citep{khandelwal2020nearest}. And we use the NAT model trained on the WMT14 De-En dataset as the pre-trained NAT model.

In the implementation, we adopt fairseq\footnote{\url{https://github.com/pytorch/fairseq}} for NMT models and faiss\footnote{\url{https://github.com/facebookresearch/faiss}} for $k$NN to reproduce the baselines as well as build our method. For all datasets, we use faiss to learn $4$k cluster centroids, and search $32$ clusters in inference.
We tune all hyper-parameters on the valid set of each task and set $n=2$, $k=8$, $\tau=10$ for all experiments. The hyper-parameter $\lambda$ is tuned for each task independently.
We analyze the memory usage of the proposed $n$-$k$NN as well as the vanilla $k$NN in Appendix~\ref{appen:exp}.

\paragraph{Baseline and Evaluation}
We denote the model trained on general tasks as Base AT/NAT model. We consider the vanilla $k$NN method as the baseline, and we denote our method as $n$-$k$NN. 
We use SacreBLEU\footnote{\url{https://github.com/mjpost/sacrebleu}}~\citep{post2018call} to measure the case-sensitive detokenized BLEU~\citep{papineni2002bleu} scores on domain adaptation tasks, and case-sensitive tokenized BLEU scores on general translation tasks, to keep consistent with settings in previous works.

\subsection{Main Results}

The results of our method on domain adaptation and general translation are listed in Table~\ref{tab:domain_bleu_results} and Table~\ref{tab:general_bleu_results} respectively.
From Table~\ref{tab:domain_bleu_results}, the proposed $n$-$k$NN method consistently outperforms the vanilla $k$NN method on different domains and models. For AT models, $n$-$k$NN brings $1.03$ BLEU score improvements on average. And for NAT models, the vanilla $k$NN fails to provide better results on all domains, indicating that the values cannot be accurately retrieved by the unigram hidden representations. On the contrary, the proposed $n$-$k$NN method brings significant improvements on all domains with $2.76$ BLEU scores on average, showing that the $n$-gram representation as well as the equipped two-pass decoding algorithm can successfully store and retrieval informative context representations.

As for the results on general translation tasks, the proposed $n$-$k$NN method also brings consistent improvements over the vanilla $k$NN method. Specifically, on the NAT model, $n$-$k$NN brings $0.48$ and $1.10$ improvements of BLEU scores on WMT14 De-En and WMT16 En-Ro tasks respectively. Note that the proposed method is model agnostic, and expected to bring improvements when applied on other NAT models.

\subsection{Ablation Study}

\begin{table}[tb]
\small
\centering
\caption{The ablation study on AT and NAT models regarding the proposed $n$-$k$NN and corresponding decoding algorithm. 
}
\vspace{-10pt}
\begin{tabular}{l|c}
\toprule
 \multicolumn{1}{l|}{\textbf{Models}} & \multicolumn{1}{c}{\textbf{IT}}  \\
\midrule
(1): Base AT  & $37.97$ \\
(2): (1) + $k$NN  & $44.54$ \\
(3): (2) + Update Beam   & $45.63$  \\
(4): (1) + $n$-$k$NN w/o Update Beam & $44.27$  \\
(5):  $n$-$k$NN + Update Beam   & $\textbf{46.42}$  \\
\midrule
(1): Base NAT   & $30.60$  \\
(2): (1) + $k$NN  & $30.59$  \\
(3): (1) + $n$-$k$NN & $31.21$  \\
(4): (1) + Two-Pass  & $31.19$  \\
(5): (2) + Two-Pass  & $31.39$  \\
(6):  $n$-$k$NN + Two-Pass & $\textbf{31.43}$   \\
\bottomrule
\end{tabular}
\label{tab:nat-ablation}
\vspace{-5pt}
\end{table}

\begin{figure}[tb]
\centering
\includegraphics[width=0.8\linewidth]{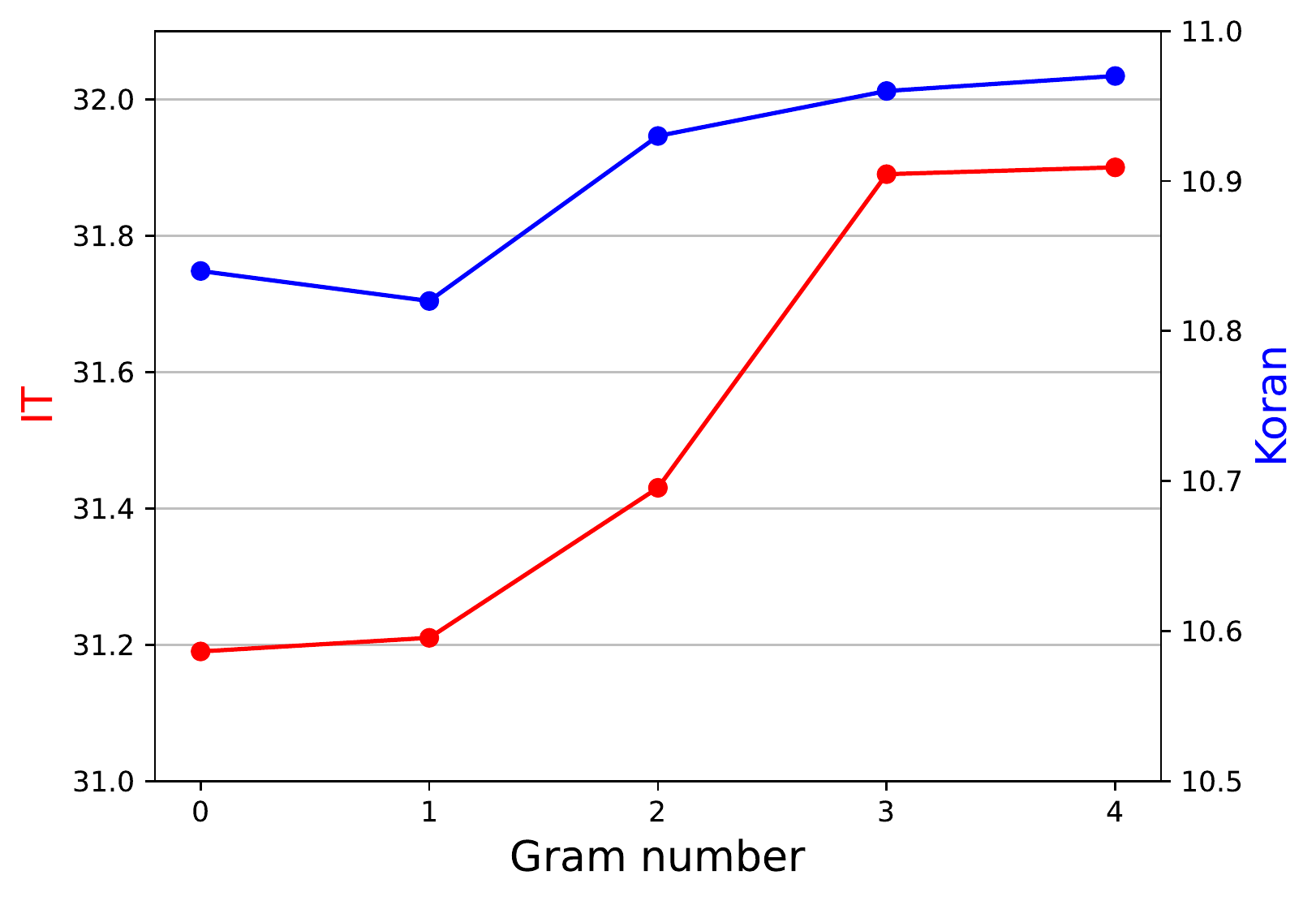}
\vspace{-5pt}
\caption{The performance of $n$-$k$NN NAT with different $n$-gram numbers on IT and Koran domain.}
\vspace{-5pt}
\label{fig:n-gram-study}
\end{figure}

\begin{table}[tb]
\small
\centering
\caption{The top-$5$ accuracy of the datastore and the ratio of repetitive translation error of different models. Evaluation is conducted on the NAT model and IWSLT14 De-En dataset.}
\vspace{-10pt}
\begin{tabular}{l|cccc}
\toprule
 \multicolumn{1}{c|}{Models}   & \multicolumn{1}{c}{\textbf{Top-$5$ Acc}}& \multicolumn{1}{c}{\textbf{Rep. Ratio}}    \\
\midrule
Base NAT & $23.24\%$ & $4.37\%$     \\
$+$ $k$NN & $23.17\%$ & $4.39\%$    \\
$+$\textbf{$n$-$k$NN} & $\textbf{31.20\%}$& $\textbf{3.46\%}$   \\
\bottomrule
\end{tabular}
\label{tab:analyse}
\vspace{-10pt}
\end{table}

\begin{table*}[tb]
\small
\centering
\caption{Some cases of the translation results of different models on the IWSLT14 De-En for AT and the WMT14 De-En Medical task for NAT. 
The bold italics represent the repetitive words in translation results.}
\vspace{-10pt}
\begin{tabular}{r|l}
\toprule
\specialrule{0em}{1pt}{1pt}
\multirow{1}{*}{Source:} & als kind zeichnete ich ständig comicbücher und so. \\
\specialrule{0em}{1pt}{1pt}
\cline{2-1}
\specialrule{0em}{1pt}{1pt}
\multirow{1}{*}{Target:} & and i was, as a kid, constantly drawing comic books, and so on. \\
\specialrule{0em}{1pt}{1pt}
\cline{2-1}
\specialrule{0em}{1pt}{1pt}
\multirow{1}{*}{Basic AT:} & as a kid, i was constantly drawing comic books \textbf{\textit{and things}}. \\
\specialrule{0em}{1pt}{1pt}
\cline{2-1}
\specialrule{0em}{1pt}{1pt}
\multirow{1}{*}{$k$NN-AT:} & as a kid, i was cartooning comic books all the time, \textbf{\textit{so on}} . \\
\specialrule{0em}{1pt}{1pt}
\cline{2-1}
\specialrule{0em}{1pt}{1pt}
\multirow{1}{*}{$n$-$k$NN-AT:} & as a kid, i was constantly drawing comic books, and so on. \\
\hline
\specialrule{0em}{1pt}{1pt}
\multirow{2}{*}{Source:} & Vorne sind die vielfach gerüschten und mit Stickereien versehenen Kleider geschlossen ,  \\
& oroft mit einem Schleier , hinten hingegen gehen sie den Rücken hinab und enden oft in einer Schleppe .\\
\specialrule{0em}{1pt}{1pt}
\cline{2-1}
\specialrule{0em}{1pt}{1pt}
\multirow{2}{*}{Target:} & At the front , dresses with numerous ruffles and embroidery are in , often with a veil , \\
&  while to the rear they run down the back and often end in a train .\\

\specialrule{0em}{1pt}{1pt}
\cline{2-1}
\specialrule{0em}{1pt}{1pt}
\multirow{2}{*}{Basic NAT:} & In the front are closed clothes , which are often \textbf{\textit {and and}} fitted with embroidery , \\
& often with a veil , but at the back they go down their back and often end in a tol .\\
\specialrule{0em}{1pt}{1pt}
\cline{2-1}
\specialrule{0em}{1pt}{1pt}
\multirow{2}{*}{$k$NN-NAT:} & In the front are closed\textbf{\textit{ of of often of and of }}with embroidery , \\
& often with a veil , but at the back they go down their back and often end in a to.\\
\specialrule{0em}{1pt}{1pt}
\cline{2-1}
\specialrule{0em}{1pt}{1pt}
\multirow{2}{*}{$n$-$k$NN-NAT:} & In the front are closed clothes ,which are often \textit{armed and} fitted with embroidery ,  \\
& often with a veil , but at the back they go down their backs and often end in a tol . \\
\specialrule{0em}{1pt}{1pt}
\bottomrule
\end{tabular}
\label{tab:case_study}
\vspace{-5pt}
\end{table*}

We conduct various ablation studies in this subsection to verify the proposed components including $n$-gram datastore creation and corresponding decoding algorithms for AT and NAT.

The ablation study results are listed in Table~\ref{tab:nat-ablation}. For AT model, we can find that if only using the keys related to the current timestep, the $n$-gram performance is actually inferior to the unigram counterpart (setting (4) verse (2)). But when updating beam scores using the keys before the current timestep, the information contained in retrieved $n$-grams can be fully utilized, and achieve better performance (setting (5) verse (4)). 

As for the ablation study of NAT model, according to Table~\ref{tab:nat-ablation}, with the vanilla one-pass decoding algorithm, the unigram method fails to provide useful information with ambiguous representations, and thus performs even worse than the raw model.
The proposed $n$-$k$NN method performs better than the baseline 
as $n$-gram representations can help alleviate the errors that unigram makes.
When equipped with the proposed two-pass decoding, both vanilla $k$NN and $n$-$k$NN bring improvements over the baseline, showing the importance of good representations. And the proposed $n$-gram retrieval and two-pass decoding are complementary to each other for achieving good results.

\paragraph{Ablation on Gram Number $n$}
We set $n=2$ for the main results to show the generality of our method, which can be easily scaled to larger numbers. We study the performance of $n$-$k$NN on the NAT model with different $n$-gram numbers in Figure~\ref{fig:n-gram-study}.
Setting $n=1$, i.e., vanilla $k$NN, does not provide consistent improvements. With $n \geq 2$, $n$-$k$NN significantly outperforms $k$NN, and the performance is better with larger $n$s. The improvements over $n>3$ become marginal, and we conjecture the reason is that the extra co-occurrence information and the noise provided by large $n$-grams cancel out each other.

\subsection{Analyses}
We provide fine-grained analyses of the proposed method in this subsection. 
NAT models can achieve high decoding speed but with the problem of repetitive translations, as stated in Section~\ref{sec:intro}.
In Table~\ref{tab:analyse}, we calculate the ratio of repetitive translation errors and the top-$5$ prediction accuracy to measure the quality of the datastore built by different methods. 
For the repetitive translation error, $n$-$k$NN is able to reduce the repetition rate by $\sim 1\%$ in a non-parametric way.
In addition, we calculate the top-$5$ prediction accuracy to measure the quality of the datastore built by different methods. Specifically, given $5$ retrieved neighbors from the datastore, it is considered accurate if the golden target token appears in it.
As a result,
the proposed $n$-$k$NN method achieves higher accuracy than baselines, indicating the better quality of the datastore.
In addition, on the AT model, we verify the better disambiguation ability carried by $n$-$k$NN over the vanilla $k$NN through the Word Sense Disambiguation~(WSD) test~\citep{rios-gonzales-etal-2017-improving}, as shown in Appendix~\ref{appen:contraWSD}.

\subsection{Case Study}

We provide case studies in this subsection to intuitively show the improvements brought by the proposed $n$-$k$NN method. Table~\ref{tab:case_study} shows cases from the general IWSLT14 De-En domain and the Medical domain, on AT and NAT models respectively. We can find that in both cases, the proposed method corrects the ambiguous translation error for AT and the repetitive translation error of NAT, i.e., from ``and things'' to ``and so on'', and from ``and and '' to ``armed and'', showing the benefits brought by utilizing more distinguishable context representation of $n$-grams.

\section{Conclusion}
In this paper, we propose a novel $n$-gram $k$ nearest neighbor method which can be utilized on both autoregressive and non-autoregressive models. Instead of taking the unigram context representation as the key and query of the datastore, we use the consecutive $n$-gram representations as well as their corresponding target tokens as the key and value to construct the datastore. In inference, to fully utilize the information in the retrieved $n$-gram values, we design tailored decoding algorithms for AT and NAT models considering their different decoding manner, i.e., the $n$-gram augmented beam search decoding for AT, and the two-pass decoding for NAT. In experiments, the proposed $n$-$k$NN method consistently outperforms the vanilla $k$NN on both general and domain adaptation translation tasks, with $1.03$ and $2.67$ BLEU score improvements on average on AT and NAT models respectively.
In the future, we plan to extend our method to other scenarios where source information is crucial such as document translation, text summarization, etc. 
Limitations are discussed in Appendix~\ref{appen:limit}.

\bibliography{acl_latex}
\bibliographystyle{acl_natbib}

\newpage

\appendix
\section{Appendix}
\label{sec:appendix}

\subsection{Experiment Setup}
\label{appen:exp}

\begin{table}[tb] \small
\centering
\caption{Statistics of the multi-domain dataset.}
\vspace{-5pt}
\begin{tabular}{l|cccc}
\toprule
\multicolumn{1}{l|}{Dataset} & IT    & Medical & Koran & Laws    \\ \midrule
Train                     & $223$k & $248$k   & $18$k & $467$k \\ 
Dev            & $2$k & $2$k   & $2$k & $2$k \\ 
Test                     & $2$k & $2$k   & $2$k & $2$k \\ 
\bottomrule
\end{tabular}%
\label{table:domain_dataset}
\end{table}

\begin{table}[tb] \small
\centering
\caption{Statistics of the general dataset.}
\vspace{-5pt}
\begin{tabular}{l|ccc}
\toprule
 \multicolumn{1}{c|}{}   & \multicolumn{1}{c}{{IWSLT14}}& \multicolumn{1}{c}{{WMT14}} & \multicolumn{1}{c}{{WMT16}}    \\
  Dataset &\multicolumn{1}{c}{{De-En}}& \multicolumn{1}{c}{{De-En}} & \multicolumn{1}{c}{{En-Ro}} \\
\midrule
Train                     & $157$k & $4.5$M   & $610$k  \\ 
Dev            & $7$k  & $3$k & $2$k \\ 
Test                     & $7$k & $3$k   & $2$k  \\ 
\bottomrule
\end{tabular}%
\label{table:general_dataset}
\end{table}

\begin{table}[tb] \small
\centering
\caption{Memory usage of $k$NN and $n$-$k$NN.}
\vspace{-5pt}
\begin{tabular}{l|cccc}
\toprule
\multicolumn{1}{l|}{Memory Size} & IT    & Medical & Koran & Laws    \\ \midrule
$k$NN & $266$M & $492$M & $54$M & $1.3$G \\
$n$-$k$NN & $266$M & $492$M & $54$M & $1.3$G \\
\bottomrule
\end{tabular}%
\label{table:domain_memory}
\end{table}

\begin{table}[tb]
\small
\centering
\caption{ The inference latency of different models on the IWSLT14 De-En test set.}
\vspace{-10pt}
\begin{tabular}{l|c|c|cc}
\toprule
 \multicolumn{1}{c|}{Models}   & \multicolumn{1}{c}{\textbf{Latency}}& \multicolumn{1}{|c|}{Models} & \multicolumn{1}{c}{\textbf{Latency}}    \\
\midrule
Base AT & $245$ms & Two-Pass NAT & $45$ms    \\
$+$ $k$NN & $552$ms & $+$ $k$NN & $66$ms    \\
$+$\textbf{$n$-$k$NN} & $574$ms & $+$\textbf{$n$-$k$NN} & $75$ms  \\
\bottomrule
\end{tabular}
\label{tab:latency}
\vspace{-10pt}
\end{table}

For the considered four domains include Koran, IT, Medical and Law, their statistics are listed in Table~\ref{table:domain_dataset}. For the general dataset including IWSLT14 De-En, WMT14 De-En and WMT16 En-Ro, their statistics are listed in Table~\ref{table:general_dataset}.
We strictly follow the split of training/valid/test sets utilized in previous works.
The memory usage is listed in Table~\ref{table:domain_memory}, where we report the size of the faiss index, and we can find that $n$-$k$NN has similar memory usage as the vanilla $k$NN after the approximation of faiss.

For the model architecture, on general translation tasks, we use the \texttt{base} Transformer setting~($d_{\textrm{hidden}}=512$, $d_{\textrm{ffn}}=2048$, $n_{\textrm{layer}}=6$, $n_{\textrm{head}}=8$) for WMT tasks, and a smaller setting~($d_{\textrm{ffn}}=1024$, $n_{\textrm{head}}=4$) for the IWSLT task.

In addition,
we list the per-sentence decoding latency of different models on the test set of IWSLT14 De-En in Table~\ref{tab:latency}, tested on an NVIDIA P100 GPU.
The results
demonstrate that 
the $n$-gram retrieval has a similar decoding speed with the unigram baseline as we have discussed in Section~\ref{sec:discuss}.
It is worth noting that while both $k$NN and $n$-$k$NN models are slower than the base model, recent works~\citep{he2021efficient,wang2022efficient,alon2022neuro} that improve the efficiency of $k$NN based models can also be utilized in the proposed $n$-$k$NN model smoothly. We take the application for future work.

\subsection{Word Sense Disambiguation Test}
\label{appen:contraWSD}

\begin{table}[tb]
\small
\centering
\caption{The accuracy rate of ambiguous words test on the AT model.}
\vspace{-10pt}
\begin{tabular}{l|cccc}
\toprule
 \multicolumn{1}{c|}{Models}   & \multicolumn{1}{c}{Base AT } & \multicolumn{1}{c}{{$+$ $k$NN}} &\multicolumn{1}{c}{$+$\textbf{$n$-$k$NN}}   \\
\midrule
\textbf{Acc} & $91.34\%$  &  $91.49\%$ &  $\textbf{95.88\%}$ \\
\bottomrule
\end{tabular}
\label{tab:ambiguous-test}
\vspace{-10pt}
\end{table}

As we introduced in Section~\ref{sec:intro}, 
the same word often has different senses given different context,
and the token-level representations sometimes fail to deal with ambiguous context information. 
To demonstrate that $n$-$k$NN can alleviate this problem, we conduct the word sense disambiguation test on the ContraWSD~\citep{rios-gonzales-etal-2017-improving} dataset, which consists of sentence pairs with ambiguous German words extracted from the WMT German-English dataset. Specifically, each sentence pair has a reference translation and a set of contrastive translations, which are composed by replacing the ambiguous words of the reference translation with one of its other meanings.
As an example, we consider one German word \textit{Schlange}, which will be translated to \textit{snake} with the current context. And the word also has three ambiguous translations: \textit{serpent, line, queue}. In the contrastive translated sentences, the word \textit{snake} in the reference translation will be replaced by one of these ambiguous words, and then the source sentence will have four references, one is the golden reference and the other three are contrastive references. 
If the model assigns a higher BLEU Score to the golden reference than to all contrastive translations, it is counted as a correct decision. We use the AT models trained on the WMT14 DE-EN dataset to do the ambiguous test and list the accuracy rates in Table~\ref{tab:ambiguous-test}. The results indicate that the unigram $k$NN method performs similarly to the base model, i.e., does not promote the disambiguation ability of the model. And our method achieves better accuracy when distinguishing ambiguous words by using phrase-level information.

\subsection{Case Study}
\label{appen:case}

we provide extra case studies of the proposed $n$-$k$NN method on both AT and NAT models in Table~\ref{tab:case_studyx}. On the contrary, due to the low quality of keys and queries as well as the inaccurately retrieved values for NAT models, the vanilla $k$NN method fails to provide better results on neither general nor domain adaptation translations.

\begin{table*}[htb]
\small
\centering
\begin{tabular}{r|l}
\toprule

\specialrule{0em}{1pt}{1pt}
\multirow{1}{*}{Source:} & wenn bei Ihnen eine Erkrankung Ihres Immunsystems diagnostiziert ist . \\
\specialrule{0em}{1pt}{1pt}
\cline{2-1}
\specialrule{0em}{1pt}{1pt}
\multirow{1}{*}{Target:} & If you have been diagnosed as having a disorder of your immune system . \\
\specialrule{0em}{1pt}{1pt}
\cline{2-1}
\specialrule{0em}{1pt}{1pt}
\multirow{1}{*}{Basic AT:} & if you \textbf{\textit{are diagnosed}} with a disease of \textbf{\textit{your immune system}} . \\
\specialrule{0em}{1pt}{1pt}
\cline{2-1}
\specialrule{0em}{1pt}{1pt}
\multirow{1}{*}{$k$NN-AT:} & if you \textbf{\textit{have Wegener}} ; s \textbf{\textit{granulomatosis}} . \\
\specialrule{0em}{1pt}{1pt}
\cline{2-1}
\specialrule{0em}{1pt}{1pt}
\multirow{1}{*}{$n$-$k$NN-AT:} & if you \textbf{\textit{have been diagnosed }} with a disease of \textbf{\textit{your immune system }}.\\
\hline

\specialrule{0em}{1pt}{1pt}
\multirow{2}{*}{Source:} & NovoNorm ist ein orales Antidiabetikum , das Repaglinid enthält und das Ihrer Bauchspeicheldrüse hilft ,  \\
& mehr Insulin zu produzieren und damit Ihren Blutzucker ( Glucose ) zu senken .\\
\specialrule{0em}{1pt}{1pt}
\cline{2-1}
\specialrule{0em}{1pt}{1pt}
\multirow{2}{*}{Target:} &NovoNorm is an oral antidiabetic agent containing repaglinide  \\
& which helps your pancreas produce more insulin and thereby lower your blood sugar ( glucose ) .\\

\specialrule{0em}{1pt}{1pt}
\cline{2-1}
\specialrule{0em}{1pt}{1pt}
\multirow{2}{*}{Basic NAT:} & NovoNorm is an oral antidiabetic that contains repaglinide \\
&  and \textbf{\textit{ helps helps}}  your pancreas to produce more insulin and thus reduce blood \textbf{\textit{glucose ( glucose )}} .
\\
\specialrule{0em}{1pt}{1pt}
\cline{2-1}
\specialrule{0em}{1pt}{1pt}
\multirow{2}{*}{$k$NN-NAT:} & NovoNorm is an oral antidiabetic that contains repaglinide \\
&  and \textbf{\textit{ helps helps}}  your pancreas to produce more insulin and thus reduce blood \textbf{\textit{glucose ( glucose )}} .\\
\specialrule{0em}{1pt}{1pt}
\cline{2-1}
\specialrule{0em}{1pt}{1pt}
\multirow{2}{*}{$n$-$k$NN-NAT:} & NovoNorm is an oral antidiabetic containing repaglinide  \\
&  which \textit{helps} your pancreas produce more insulin and thereby and reduce your \textit{glucose} . \\
\specialrule{0em}{1pt}{1pt}
\bottomrule
\end{tabular}
\caption{Some cases of the translation results of different models on the IWSLT14 De-En for AT and the WMT14 De-En Medical task for NAT. 
The bold italics represent the repetitive words in translation results.}
\label{tab:case_studyx}
\end{table*}

\subsection{Limitations}
\label{appen:limit}
The main limitation of our work is akin to that of the vanilla $k$NN work, i.e., we need to load a datastore in inference which will slower the decoding speed since an additional retrieval procedure is introduced. We plan to accelerate the retrieval process by reducing the dimension of representation, using clustering methods to reduce the number of keys contained in the datastore, etc.

\end{document}